\documentclass[letterpaper,10pt,conference]{ieeeconf}

\IEEEoverridecommandlockouts

\usepackage{etoolbox}

\AtBeginDocument{\frenchspacing}

\usepackage[final,nopatch=footnote]{microtype}

\usepackage{stix}
\usepackage{roboto}
\usepackage{circledsteps}

\usepackage[noend,noline,plainruled]{algorithm2e}

\DontPrintSemicolon
\SetAlgoCaptionSeparator{.}
\SetAlCapNameFnt{\footnotesize}
\SetAlCapFnt{\mdseries\footnotesize}
\SetAlCapSkip{0.5em}
\SetAlgoInsideSkip{medskip}
\SetKwSty{textsc}

\SetKwInput{KwData}{\mdseries\textsc{data}}
\SetKwInput{KwResult}{\mdseries\textsc{result}}

\usepackage[x11names]{xcolor}
\usepackage{hyperref}

\hypersetup{
  colorlinks=true,
  citecolor=gray,
  linkcolor=gray,
  urlcolor=gray
}

\input{glyphtounicode}
\usepackage{multirow}
\usepackage{booktabs}
\usepackage{rotating}
\usepackage{graphicx}
\usepackage{import}

\BeforeBeginEnvironment{tabular*}{\footnotesize}
\AtBeginDocument{\allowdisplaybreaks[1]}
\AtBeginEnvironment{tabular}{}

\makeatletter
\g@addto@macro\@floatboxreset\centering
\makeatother

\usepackage{svg}
\usepackage{transparent}

\usepackage{pgfplots}
\pgfplotsset{compat=newest}

\usepackage{amsmath}
\usepackage{mathtools}
\usepackage{pifont}

\usepackage{annotate-equations}

\tikzset{annotate equations/arrow/.style={-, shorten <=2pt}}
\usepackage{standalone}
\tikzset{
  every node/.append style={font=\sffamily}
}
\usetikzlibrary{arrows.meta}

\usepackage{siunitx}

\newcommand{\argmin}{\mathop{\rm argmin}}

\usepackage[
  doi=false,
  isbn=false,
  url=true,
  eprint=false,
  style=ieee,
  citestyle=numeric-comp,
  datamodel=software,
  maxbibnames=99,
  natbib = true
]{biblatex}

\DeclareDelimFormat{multicitedelim}{\addcomma\,}
\DeclareDelimFormat[bib]{multicitedelim}{\addcomma\,}

\usepackage{software-biblatex}

\preto{\cite}{\unskip~}

\makeatletter
\patchcmd{\@IEEEyesnumber}
  {\stepcounter}
  {\refstepcounter}
  {}{}

\patchcmd{\@@IEEEeqnarray}
  {\stepcounter}
  {\refstepcounter}
  {}{}

\patchcmd{\@@IEEEeqnarraycr}
  {\stepcounter{IEEEsubequation}}
  {\refstepcounter{IEEEsubequation}}
  {}{}

\patchcmd{\@@IEEEeqnarraycr}
  {\stepcounter{IEEEsubequation}}
  {\refstepcounter{IEEEsubequation}}
  {}{}

\patchcmd{\@@IEEEeqnarraycr}
  {\stepcounter{IEEEequation}}
  {\refstepcounter{IEEEequation}}
  {}{}

\patchcmd{\@@IEEEeqnarraycr}
  {\stepcounter{IEEEequation}}
  {\refstepcounter{IEEEequation}}
  {}{}
\makeatother

\usepackage[capitalise]{cleveref}

\usepackage{ifthen}

\makeatletter
\newcounter{IEEE@bibentries}

\renewcommand\IEEEtriggeratref[1]{%
  \renewbibmacro{finentry}{%
    \stepcounter{IEEE@bibentries}%
    \ifthenelse{\equal{\value{IEEE@bibentries}}{#1}}
      {\finentry\@IEEEtriggercmd}
      {\finentry}%
  }%
}
\makeatother

\newcommand{\contingencygames}{%
  \texttt{Contingency\allowbreak Games.jl}}
\newcommand{\mixedmcp}{%
  \texttt{Mixed\allowbreak Complementarity\allowbreak Problems.jl}}

\usepackage{makecell}
\usepackage{xcolor}
\definecolor{grayish}{HTML}{a5a5a5}
\definecolor{gold}{HTML}{eda43c}
\definecolor{redish}{HTML}{ff5555}
\definecolor{blueish}{HTML}{4d7bad}
\definecolor{yellowish}{HTML}{d57f32}

\usepackage{soul}
\usepackage{layout}
\usepackage{nicefrac}

\title{Cascaded consensus splitting for multi-branch contingency games}
\author{Bastien Lechardoy, Pau de las Heras Molins, Thibault Lahire, Laurent Pautet,\\ David Filliat, David Fridovich-Keil, and Georgios Bakirtzis%
\thanks{Corresponding author: B. Lechardoy (bastien.lechardoy@telecom-paris.fr).}%
\thanks{B. Lechardoy, P. de las Heras Molins, L. Pautet, and G. Bakirtzis are with LTCI, Télécom Paris, Institut Polytechnique de Paris.}%
\thanks{T. Lahire is with Dassault Aviation.}%
\thanks{D. Filliat is with  Pôle Recherche, AMIAD.}%
\thanks{D. Fridovich-Keil is with The University of Texas at Austin.}%
}
\begin{document}
\maketitle
\begin{abstract}
Contingency games enable agents to anticipate and plan for other agents’ hypothetical intents by constructing trajectories with a shared prefix and intent-dependent branches. While contingency games capture intent uncertainty, existing formulations rely on a single branching time, oversimplifying interactions in which different agents’ intentions are revealed at different times. Moreover, the computational cost of such problems grows rapidly with the number of agents and intents, as all scenario-dependent best responses must be solved jointly. We introduce a multi-branch contingency architecture in which sources of intent uncertainty can be resolved progressively at different branching times, allowing the planned trajectories to adapt to the asynchronous revelation of intents. We also develop an ADMM-based solver that exploits scenario-level parallelism. Experiments on tightly coupled three-agent interactions support that the proposed architecture outperforms the conventional single-branch while achieving a lower mean receding-horizon solve time.
\end{abstract}

\section{Introduction}

\begin{figure}[!t]
    \centering
    \begingroup
        \sffamily
        \def\svgwidth{\columnwidth}
        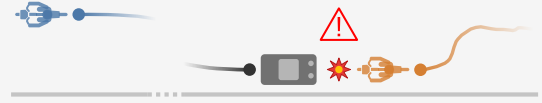
    \endgroup
    \caption{Collision avoidance scenario in which a robot must account for different behaviors of the other agents depending on possible intents. The robot does not gain information about the two uncertain agents simultaneously, and committing to a behavior too early after assuming certainty about the second agent's intention could be dangerous.}
    \label{figure:front_fig}
\end{figure}

\noindent
Autonomous agents must make decisions before the intents of other agents are fully resolved. Consider a robot interacting with a cyclist or a pedestrian. The cyclist may hesitate, continue crossing, or turn back, while the vehicle may yield or proceed (Fig.~\ref{figure:front_fig}). These possible behaviors are strategically coupled to the robot's controls, which must remain compatible with several possible futures, while retaining the ability to adapt when other agents' future behavior reveals their intents.

Contingency games provide a game-theoretic framework for planning under intent uncertainty \cite{peters:2024}. An ego agent reasons over multiple hypotheses about the latent objectives of other agents and constructs a set of intent-conditioned interaction trajectories. These trajectories share a common prefix while uncertainty remains unresolved, and diverge once other agents' behavior is expected to have revealed their intentions.
This structure delays commitment to a particular hypothesis while anticipating proper strategic responses to each possible intent.

Existing contingency game formalisms impose a single branching time across all intent hypotheses, implicitly assuming that uncertainty about every agent is resolved simultaneously. In realistic multi-agent interactions, confidence about different agents' intents may be acquired at different times. Premature commitment can be dangerous; conversely, selecting a later branching time prevents the ego agent from exploiting past observations, yielding conservative, potentially inefficient behavior.

Representing asynchronous information gain requires richer contingency structures that further exacerbate the computational limitations of contingency games. Existing methods consider multiple intent-conditioned trajectory games
coupled through shared consensus constraints, but assemble their optimality conditions into a single centralized system of equations and complementarity conditions to solve. As the planning horizon and the number of agents or intents increase, the resulting growth in problem dimension severely limits the practical deployment of such solvers.

In this paper, we introduce contingency games with multiple branching times and develop a consensus decomposition for solving them efficiently. The alternating direction method of multipliers (ADMM) \cite{boyd:2011} allows an optimization problem with shared consensus constraints to be decomposed into smaller scenario-specific subproblems. Scenario-dependent interaction games are solved in parallel following the approach of
Li et al.~\cite{li:2023}, while their local trajectories are progressively reconciled over shared segments. When uncertainty about an agent's intent is resolved, the corresponding scenario group separates, and consensus is retained only among the scenarios that remain indistinguishable (Fig.~\ref{figure:front_fig}).

In Monte Carlo collision-avoidance experiments over 100 randomized interaction instances with asynchronously revealed intents, the proposed multi-branch formulation increases the interaction success rate by \(42\%\) relative to the conventional single-branch approach, from \(66\%\) to \(94\%\). Among scenarios successfully completed by both methods, it reduced the closed-loop cost by \(15\%\) and control variation by \(26\%\), indicating more efficient and comfortable trajectories. Moreover, the average online solve time was reduced by \(73\%\), from \(1.72\) to \(0.46\,\mathrm{s}\) per receding-horizon step.

\section{Related work}

Relevant prior work concerns game-theoretic motion planning, contingency games, and distributed scenario games.

\smallskip
\noindent
\textsc{game-theoretic motion planning} \quad
Differential games model strategic interactions among agents whose states evolve according to continuous-time differential dynamics \cite{isaacs:1965}. In robotics, this framework is specialized to trajectory games that address discrete-time motion planning problems in domains such as autonomous driving \cite{Mingyu-Wang:2021,ji:2023}, multi-robot coordination \cite{le:2024,Michalis:2017} and drone racing \cite{Spica:2020, Zijian-Wang:2020}. Trajectory games are generalized Nash equilibrium problems (GNEP) \cite{basar:1982} in which robots optimize their objectives subject to constraints depending on the decisions of other agents, who themselves are modeled as optimizing objectives that depend upon robot actions. The resulting equilibrium problem is strongly coupled and becomes very hard to solve in high dimensions \cite{facchinei:2010}. Computational complexity is further exacerbated when uncertainty is incorporated into the model, e.g. in human-robot interactions \cite{Laine:2021,tian:2022}. In this work, we improve the tractability of contingency games, a class of trajectory games that considers parametric uncertainty over agents' intent.

\smallskip
\noindent
\textsc{contingency games} \quad
Handling uncertainty about other agents' intent is a key challenge in multi-agent strategic interaction \cite{hu:2023}. Certainty-equivalence frameworks assume the most probable hypothesis is true, producing efficient but hazardous solutions \cite{liu-certainty-equivalent:2023}, while fixed-uncertainty methods optimize trajectories under a fixed belief distribution over hypotheses \cite{Laine:2021}. Peters et al. \cite{peters:2024} proposed contingency games, which encode a simplified model for an ego agent's uncertainty over other agents' intentions that can be expected to resolve in the future. Contingency games delay commitment to a particular hypothesis by constructing intent-conditioned trajectories that share a common prefix until a point of sufficient certainty is expected to be reached. However, this formulation imposes a single ``branching time,'' implicitly assuming that the robot simultaneously resolves its uncertainty about every other agent's intent. In practical human-robot interactions, however, it is normal for some (e.g., nearby) agents' intents to be revealed before others. We extend contingency games to operate with multiple branching times in order to account for this type of asynchronous resolution of intent uncertainty. This change can substantially increase the size of the resulting GNEP, necessitating improvements in solver design that exploit partial consensus-splitting ADMM.

\smallskip
\noindent
\textsc{distributed scenario games} \quad
Uncertainty-aware trajectory games grow in size by accounting for more agents and intents, making them challenging to solve. To overcome this scaling challenge, we build upon the splitting method developed by Li et al. \cite{li:2023}, which uses the alternating direction method of multipliers (ADMM) \cite{boyd:2011} to employ parallel computation to solve large-scale parameterized games. For multiple branching times, we introduce a cascaded partial-consensus ADMM decomposition that enforces agreement on trajectory prefixes within progressively refined groups of indistinguishable scenarios, enabling efficient, parallel solution of scenario-dependent subproblems.

\section{Preliminaries}
\label{sec:preliminaries}

This section introduces trajectory games and formulates the single-branch contingency game.

\smallskip
\noindent
\textsc{notation}\quad
For $n$ in $\mathbb{N}_{>0}$, define $[n]\coloneqq\{1,\ldots,n\}$. A game comprises $N$ agents indexed by $i\in[N]$. Agent indices are written as superscripts and $-i$ denotes all agents other than $i$. For any agent-indexed quantity $z^i$, omitting the superscript denotes
its stacked all-agent counterpart $z\coloneqq(z^1,\ldots,z^N)$.

\subsection{Trajectory games}

Consider a finite planning horizon $T_{\mathrm{plan}}$. The trajectory of agent $i$ is the collection of states $x^i$ and controls $u^i$ over the horizon, denoted by
\[
    \tau^i\coloneqq\left\{(x_t^i,u_t^i)\right\}_{t=0}^{T_{\mathrm{plan}}-1},
\]
while $\tau^{-i}$ collects the trajectories of all other agents.
Each agent minimizes an individual cost $J^i(\tau^i,\tau^{-i})\in\mathbb R$
subject to coupled constraints $K^i(\tau^i,\tau^{-i})\in\mathbb R^{k^i}$, where $k^i$ is the number of constraints applied to agent $i$. The best-response problem of an agent $i$ is

\smallskip
\begin{equation}
\label{eq:trajectory_game}
\begin{aligned}
    \tau^{i,\star}
    \in
    \argmin_{\tau^i}
    \quad&
    \eqnmarkbox[purple]{cost}{
        J^i(\tau^i,\tau^{-i})
    }
    \\
    \textup{subject to}
    \quad&
    \eqnmarkbox[olive]{constraint}{
        K^i(\tau^i,\tau^{-i})\geq 0
    }.
\end{aligned}
\end{equation}

\annotate[yshift=+0.5em]{above,right}
{cost}{individual cost}
\annotate[yshift=-0.5em]{below,right}
{constraint}{private constraints}

\bigskip 

A trajectory profile $\tau^\star=(\tau^{1,\star},\ldots,\tau^{N,\star})$ is a generalized Nash equilibrium (GNE) if each $\tau^{i,\star}$ solves the best-response problem in \eqref{eq:trajectory_game} given $\tau^{-i,\star}$, for every $i\in[N]$ simultaneously.

\subsection{Single-branch contingency games}
\label{sec:single_branch_contingency}

Contingency games extend trajectory games by introducing one-sided
uncertainty about the latent intents of non-ego agents
\cite{peters:2024}. In this setting, we fix $i$ as the ego agent index. For each non-ego agent $j\in[N]\setminus\{i\}$, let $\theta^j$ denote its latent intent among a set of hypothetical possibilities $\mathcal I^j$. A complete scenario $\theta$ specifies one intent for each non-ego agent as 
\[
    \theta
    \coloneqq
    \left(\theta^j\right)_{j\in[N]\setminus\{i\}}
    \in
    \Theta
    \coloneqq
    \prod_{j\in[N]\setminus\{i\}}\mathcal I^j.
\]

\noindent 
For a fixed scenario $\theta\in \Theta$, the trajectory game in
\eqref{eq:trajectory_game} is instantiated by assigning intent
$\theta^j$ to each non-ego agent $j$. In particular, agent $j$
evaluates its intent-dependent objective $J^j(\tau^j,\tau^{-j};\theta^j)$.
The robot maintains a joint probability mass function $b\colon \Theta\rightarrow[0,1]$ over these scenarios. Each possibility induces a scenario-dependent equilibrium trajectory profile $\tau_\theta\coloneqq (\tau_\theta^i,\tau_\theta^{-i})$. However, the key idea is that the ego agent should restrict its plan $\tau_\Theta^i$ so that all profiles share a common prefix, in order to account for the fact that it will not resolve uncertainty over $\theta$ until some time in the future.
Let $t_b$ denote the estimated time at which all intent uncertainty is resolved, denote by $P_b^u$ a projection matrix  that extracts the sequence of controls from the trajectory preceding $t_b$, and let variable $c^i$ represent this common control prefix. The ego jointly optimizes its scenario-dependent trajectories according to

\smallskip
\begin{subequations}
\label{eq:single_branch_game}
\begin{align}
    \tau_\Theta^{i,\star}
    \in
    \argmin_{\tau_\Theta^i}
    \quad&
    \eqnmarkbox[brown]{average}{
        \sum_{\theta\in\Theta}b(\theta)
    }
    J^i(\tau_\theta^i,\tau_\theta^{-i}) &&
    \label{eq:single_branch_ego}
    \\
    \textup{subject to}
    \quad&
    K^i(\tau_\theta^i,\tau_\theta^{-i})\geq 0,
    \qquad
    && \forall\theta\in\Theta,
    \label{eq:single_branch_private}
    \\
    &
    \eqnmarkbox[olive]{consensus_constraints}{
        P_b^u \tau_\theta^i=c^i
    },
    \qquad
    && \forall\theta\in\Theta.
    \label{eq:single_branch_consensus}
\end{align}
\end{subequations}

\annotate[yshift=+0.5em]{above,right}
{average}{belief-weighted cost}
\annotate[yshift=-0.5em]{below,right}
{consensus_constraints}{shared-prefix constraint}
\medskip

\noindent For each scenario $\theta$ in $\Theta$, every non-ego agent $j$ in $[N]\setminus\{i\}$ simultaneously solves its corresponding scenario-dependent optimization problem

\smallskip
\begin{equation}
\label{eq:single_branch_others}
\begin{aligned}
    \tau_\theta^{j,\star}
    \in
    \argmin_{\tau_{\theta}^j}
    \quad&
    J^{j}\!\left(
        \tau_\theta^j,
        \tau_\theta^{-j},
        \eqnmarkbox[brown]{theta}{\theta^j}
    \right)
    \\
    \textup{subject to}
    \quad&
    K^{j}\!\left(
        \tau_\theta^j,
        \tau_\theta^{-j}
    \right) \geq 0,
    \qquad
    \forall \theta\in\Theta.
\end{aligned}
\end{equation}

\annotate[yshift=+0.5em]{above,right}{theta}{intent parameter}

Constraint \eqref{eq:single_branch_consensus} forces all scenario-dependent ego trajectories to share the same prefix $c^i$ before $t_b$. After $t_b$, the trajectories may diverge. This single branching time formulation assumes that all intent uncertainty resolves simultaneously. Our main contribution is a formalism that relaxes this assumption by introducing multiple branching times, at which the set of plausible scenarios is progressively partitioned as each source of intent uncertainty resolves.

\section{Multi-branch contingency games}

\begin{figure*}[t]
    \centering
    \vspace*{10pt}
    \begingroup
        \sffamily
        \def\svgwidth{0.95\textwidth}
        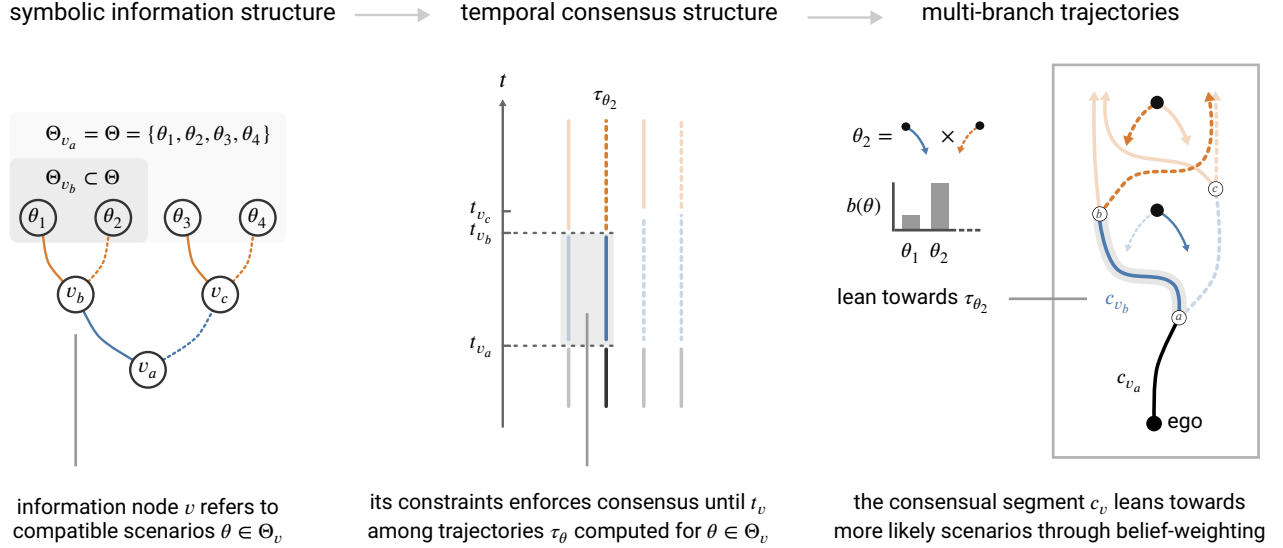
    \endgroup
    \caption{Translating asynchronous structure into consensus constraints yields multi-branch trajectories. }
    \label{fig:solution}
\end{figure*}

We extend contingency games to model interactions in which several sources of intent uncertainty are resolved asynchronously (Fig.~\ref{fig:solution}). The belief distribution and branching times are treated as fixed estimates within each open-loop planning invocation.

\subsection{Logical contingency tree and temporal realization}
\label{sec:prebranching_tree}

With multi-branching, we must distinguish between informational and temporal abstractions. The logical information tree groups scenarios in nodes according to the shared history of intent resolution. Branching times then specify for how long the trajectories associated with each group of scenarios must share a common control prefix.

\smallskip\noindent
\textsc{logical information tree} \quad
We assume that the non-ego agents' intents are revealed sequentially according to a fixed ordering. We re-index the $N-1$ non-ego agents by expected order of intent resolution. Index $q\in[N-1]$ denotes the agent resolved at level $q$. The logical evolution of accessible information is represented as a rooted tree (cf. Fig.~\ref{fig:solution}), which defines a set of internal nodes $v \in \mathcal V$. The subsets $\mathcal V_q$ contain the nodes immediately preceding the resolution of agent $q$'s intent.  Each node $v\in\mathcal V_q$
represents a distinct resolved-intent history $h_v=(\theta^1,\cdots,\theta^{q-1})$ recording the intent values resolved along the path from the root to $v$. We denote by $\Theta_v$ the set of scenarios that are consistent with that history. The root node precedes any revelation, and thus concerns all scenarios in $\Theta$. At each node $v\in\mathcal V_q$, the possible values of the intent $\theta^q$ partition $\Theta_v$ among the children of $v$. Hence, if $\operatorname{ch}(v)$ denotes the set of children of $v$, and $\biguplus$ a disjoint union,
\begin{equation}
\label{eq:tree_partitions}
\mathcal V
=
\biguplus_{q\in[N-1]}\mathcal V_q,
\qquad
\Theta_v
=
\biguplus_{w\in\operatorname{ch}(v)}\Theta_w.
\end{equation}

\noindent For each $v\in\mathcal V_q$, let $\beta_v \coloneqq \sum_{\theta\in\Theta_v} b(\theta)$ denote the probability of reaching node $v$. For \(\beta_v>0\), the marginal belief over agent $q$’s intent, conditional on the history $h_v$, is denoted by $b_v^q$ and defined as
\begin{equation}
\label{eq:node_beliefs}
\begin{aligned}
b_v^q(\eta) 
\coloneqq
\frac{1}{\beta_v}
\sum_{\substack{\theta\in\Theta_v\\\theta^q=\eta}}
b(\theta),
\quad \eta\in\mathcal I^q.
\end{aligned}
\end{equation}

Qualitatively, this structure represents a cascade of information gains. Each intent resolution progressively narrows the set of scenarios still consistent with the intent history. At node $v\in\mathcal{V}_q$, agent $q$'s intent might be revealed to be $\eta\in\mathcal I^q$ with a belief conditioned by intent history $b_v^q(\eta)$. This information structure determines which scenarios in $\Theta_v$ must prescribe the same controls until agent $q$'s intent is resolved.

\smallskip\noindent
\textsc{branching time structure} \quad 
The information tree above determines which trajectories must remain in consensus, while the node-dependent branching times specify for how long. An agent $q$ does not uniquely determine the branching time. Different intent histories can yield different future interactions where intents become distinguishable at different times. Each node $v\in\mathcal V_q$ is associated with its own branching time $t_v$ for $q\in[N-1]$. Let the projection matrix $P^u_v$ extract the ego robot's control sequence up to $t_v$, and let $c_v^i$ denote the robot's shared sequence of controls associated with node $v$. Together, the logical information tree imposes the following constraints on the ego agent's planned actions
\begin{equation}
\label{eq:tree_consensus}
    P^u_v\tau_\theta^i=c_v^i,
    \qquad
    \forall v\in\mathcal V,~
    \forall\theta\in\Theta_v.
\end{equation}

For chronological consistency, the branching times increase along each root-to-leaf path with $t_w>t_v$ for every child node $w$ of $v$. Since $\Theta_w\subseteq\Theta_v$, the scenarios reaching
a child $w$ satisfy both the consensus constraint inherited from
$v$ and the additional constraint associated with $w$. This process repeats recursively, producing a cascade of shared trajectory segments. The resulting control prefixes are therefore nested. All trajectories share the same control root, after which progressively smaller scenario groups remain in consensus until their respective branching times.

\subsection{Multi-branch contingency formulation}
\label{sec:multibranch_game}

We collect the ego consensus prefixes associated with all nodes~$v$ in $\mathcal V$ as
\[
    c_{\mathcal V}^i
    \coloneqq
    \left(c_v^i\right)_{v\in\mathcal V}.
\]
Replacing the single-prefix constraint in 
\eqref{eq:single_branch_consensus} with the tree-structured consensus constraints in \eqref{eq:tree_consensus} yields the multi-branch stacked ego problem
\begin{subequations}
\label{eq:multibranch_game}
\begin{align}
\left(
    \tau_\Theta^{i,\star},
    c_{\mathcal V}^{i,\star}
\right)
\in&
\argmin_{\tau_\Theta^i,c_{\mathcal V}^i}
\quad
\sum_{\theta\in\Theta}
b(\theta)
J^i\!\left(
    \tau_\theta^i,
    \tau_\theta^{-i}
\right)
\label{eq:multibranch_ego}
\\
\textup{subject to}\quad&
K^i\!\left(
    \tau_\theta^i,
    \tau_\theta^{-i}
\right)\geq 0,
\quad
\forall\theta\in\Theta,
\label{eq:multibranch_private}
\\
&
\llap{%
    \textcolor{olive!85}{%
        \sffamily\footnotesize
        \shortstack[r]{%
            control-prefix\\[-0.35ex]
            consensus%
        }%
    }%
    \hspace{0.4em}%
    \textcolor{olive}{%
        \raisebox{0.45ex}{\rule{0.8em}{0.4pt}}%
    }%
    \hspace{0.15em}%
}%
\eqnmarkbox[olive]{tree-consensus}{
    P_v^u\tau_\theta^i=c_v^i
},
\quad
\forall v\in\mathcal V,\;
\forall\theta\in\Theta_v.
\label{eq:multibranch_consensus}
\end{align}
\end{subequations}

The scenario-dependent optimization problems of the non-ego agents remain unchanged from problem \eqref{eq:single_branch_others}. The extension modifies only the consensus structure imposed on the ego's stacked problem. In \eqref{eq:single_branch_consensus}, all scenario-specific responses must remain identical until the single branching time. In the cascaded version, uncertainty is resolved progressively.

\section{Partial-consensus splitting ADMM} 

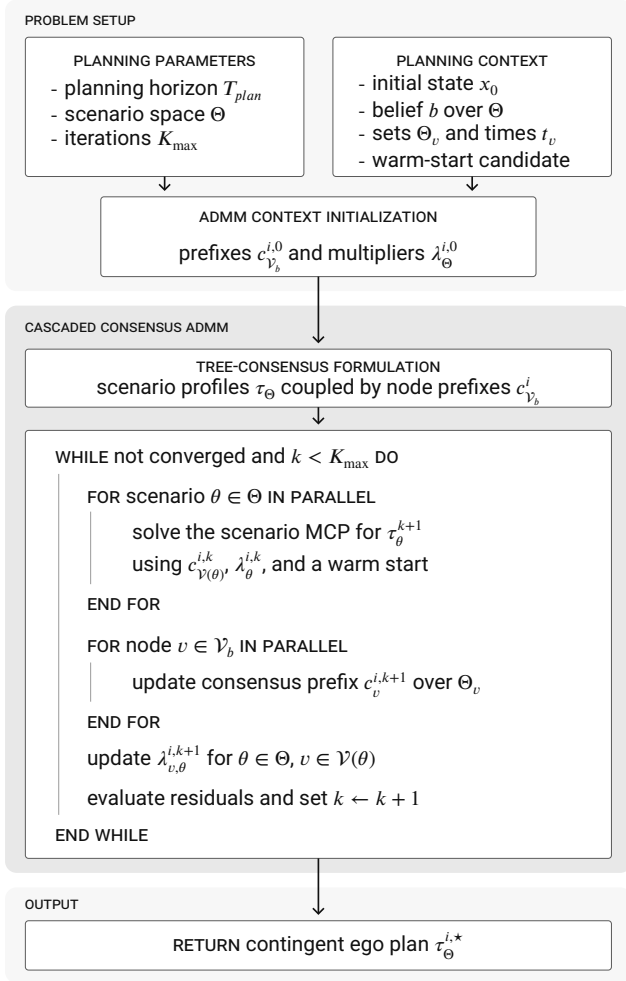
\begin{figure}[!t]
  \centering
  \begingroup%
\sffamily%
\ifcsname sansmath\endcsname\sansmath\fi%
\definecolor{oladmmPanel}{HTML}{F7F7F7}%
\definecolor{oladmmSplitPanel}{HTML}{E8E8E8}%
\definecolor{oladmmRule}{HTML}{5A5A5A}%
\definecolor{oladmmGuide}{HTML}{9A9A9A}%
\definecolor{oladmmText}{HTML}{111111}%
\providecommand{\oladmmBodyFont}{}%
\renewcommand{\oladmmBodyFont}{\sffamily\fontsize{14.4}{16.3}\selectfont}%
\providecommand{\oladmmTitleFont}{}%
\renewcommand{\oladmmTitleFont}{\sffamily\fontsize{12.6}{14.1}\selectfont\scshape}%
\providecommand{\oladmmBandFont}{}%
\renewcommand{\oladmmBandFont}{\sffamily\fontsize{11.3}{12.8}\selectfont\scshape}%
\providecommand{\kw}[1]{\textsc{#1}}%
\renewcommand{\kw}[1]{\textsc{#1}}%
\begin{tikzpicture}[x=1pt,y=-1pt]
  \tikzset{%
    panel/.style={draw=none, rounded corners=8pt},%
    block/.style={draw=oladmmRule, fill=white, rounded corners=2.2pt, line width=0.50pt},%
    flowarrow/.style={-Straight Barb, line width=1.15pt, draw=oladmmText},%
    guideline/.style={draw=oladmmGuide, line width=0.40pt, line cap=butt},%
    bodytext/.style={font=\oladmmBodyFont, text=oladmmText},%
    titletext/.style={font=\oladmmTitleFont, text=oladmmText},%
    bandtext/.style={font=\oladmmBandFont, text=oladmmText}%
  };%
  \path[panel, fill=oladmmPanel]      (4,4)   rectangle (436,205);
  \path[panel, fill=oladmmSplitPanel] (4,215) rectangle (436,606);
  \path[panel, fill=oladmmPanel]      (4,616) rectangle (436,682);

  \node[bandtext, anchor=west] at (14,18)  {problem setup};
  \node[bandtext, anchor=west] at (14,230) {cascaded consensus admm};
  \node[bandtext, anchor=west] at (14,628) {output};

  \draw[block] (18,30) rectangle (210,126);
  \node[titletext] at (114,46) {planning parameters};
  \node[bodytext, anchor=west] at (32,67)  {\textnormal{-}\hspace{0.33em}planning horizon $T_{plan}$};
  \node[bodytext, anchor=west] at (32,84)  {\textnormal{-}\hspace{0.33em}scenario space $\Theta$};
  \node[bodytext, anchor=west] at (32,101) {\textnormal{-}\hspace{0.33em}iterations $K_{\max}$};

  \draw[block] (230,30) rectangle (422,126);
  \node[titletext] at (326,46) {planning context};
  \node[bodytext, anchor=west] at (244,63)  {\textnormal{-}\hspace{0.33em}initial state $x_0$};
  \node[bodytext, anchor=west] at (244,80)  {\textnormal{-}\hspace{0.33em}belief $b$ over $\Theta$};
  \node[bodytext, anchor=west] at (244,97)  {\textnormal{-}\hspace{0.33em}sets $\Theta_v$ and times $t_v$ };
  \node[bodytext, anchor=west] at (244,114) {\textnormal{-}\hspace{0.33em}warm-start candidate};

  \draw[block] (70,140) rectangle (370,195);
  \node[titletext] at (220,154) {admm context initialization};
  \node[bodytext] at (220,181)
    {prefixes $c_{\mathcal V_b}^{i,0}$ and multipliers $\lambda_{\Theta}^{i,0}$};

  \draw[block] (18,245) rectangle (422,285);
  \node[titletext] at (220,257) {tree-consensus formulation};
  \node[bodytext] at (220,274)
    {scenario profiles $\tau_\Theta$ coupled by node prefixes $c_{\mathcal V_b}^{i}$};

  \draw[block] (18,301) rectangle (422,596);

  \draw[guideline] (41,331) -- (41,566);
  \draw[guideline] (63,357) -- (63,407);
  \draw[guideline] (63,461) -- (63,488);

  \node[bodytext, anchor=west] at (35,320)
    {\kw{while} not converged and $k<K_{\max}$ \kw{do}};

  \node[bodytext, anchor=west] at (57,346)
    {\kw{for} scenario $\theta\in\Theta$ \kw{in parallel}};
  \node[bodytext, anchor=west] at (88,372)
    {solve the scenario MCP for $\tau_\theta^{k+1}$};
  \node[bodytext, anchor=west] at (88,395)
    {using $c_{\mathcal V(\theta)}^{i,k}$, $\lambda_\theta^{i,k}$, and a warm start};
  \node[bodytext, anchor=west] at (57,421) {\kw{end for}};

  \node[bodytext, anchor=west] at (57,450)
    {\kw{for} node $v\in\mathcal V_b$ \kw{in parallel}};
  \node[bodytext, anchor=west] at (88,476)
    {update consensus prefix $c_v^{i,k+1}$ over $\Theta_v$};
  \node[bodytext, anchor=west] at (57,502) {\kw{end for}};

  \node[bodytext, anchor=west] at (57,528)
    {update $\lambda_{v,\theta}^{i,k+1}$ for $\theta\in\Theta$, $v\in\mathcal V(\theta)$};
  \node[bodytext, anchor=west] at (57,554)
  {evaluate residuals and set $k\gets k+1$};
  \node[bodytext, anchor=west] at (35,580) {\kw{end while}};

  \draw[block] (18,639) rectangle (422,673);
  \node[bodytext] at (220,656)
    {\kw{return} contingent ego plan $\tau_{\Theta}^{i,\star}$};

  \draw[flowarrow] (114,126) -- (114,136);
  \draw[flowarrow] (326,126) -- (326,136);
  \draw[flowarrow] (220,195) -- (220,241);
  \draw[flowarrow] (220,285) -- (220,297);
  \draw[flowarrow] (220,596) -- (220,635);
\end{tikzpicture}%
\endgroup%
\ignorespaces%
  \caption{Control flow of the multi-branch ADMM solver.}
  \label{figure:pseudo_code}
\end{figure}

In this section, we adapt scenario-based ADMM \cite{li:2023} to the multi-branch structure of contingency games. The goal is to preserve cascaded contingent planning while enabling parallel solving of optimization subproblems coupled through shared prefixes. The belief distribution and branching times are fixed within each open-loop solve. ADMM alternates between parallel scenario updates, conditional consensus updates, and multiplier updates (Fig.~\ref{figure:pseudo_code}).

\subsection{Scenario decomposition}

The consensus constraints in
\eqref{eq:multibranch_consensus} couple the otherwise independent scenario-dependent trajectory problems. For each complete scenario~$\theta$ in $\Theta$, define
\[
    \mathcal V(\theta)
    \coloneqq
    \left\{
        v\in\mathcal V
        \,\middle|\,
        \theta\in\Theta_v
    \right\},
\] as the set of nodes encountered from the root to the corresponding leaf of the logical information tree. For every $v \in \mathcal V(\theta)$, define the consensus residual
\begin{equation}
\label{eq:tree_consensus_residual}
    \delta_{v,\theta}^i
    \coloneqq
    P^u_v\tau_\theta^i-c_v^i.
\end{equation}
\noindent The scenario decomposition is obtained by dualizing the consensus constraints above. Their violation values are incorporated into an augmented Lagrangian through a linear term weighted by a Lagrange multiplier and a quadratic penalty controlled by parameter \(\rho\). The multipliers are  iteratively updated with ADMM, progressively driving the local scenario solutions toward consensus.

For each scenario~$\theta$, we collect the consensus prefixes and
multipliers encountered along its tree path as respectively
\[
    c_{\mathcal V(\theta)}^i
    \coloneqq
    \left\{
        c_v^i
    \right\}_{v\in\mathcal V(\theta)}
    \quad \text{and} \quad
    \lambda_\theta^i
    \coloneqq
    \left\{
        \lambda_{v,\theta}^i
    \right\}_{v\in\mathcal V(\theta)}.
\]
We further define the set of all multipliers as
\[
    \lambda_\Theta^i
    \coloneqq
    \left\{
        \lambda_\theta^i
    \right\}_{\theta\in\Theta}.
\]

\noindent In the following, recall that omitting the agent superscript denotes the corresponding all-agent trajectory profile, and define
\[
    \tau_\theta
    \coloneqq
    \left(
        \tau_\theta^i,
        \tau_\theta^{-i}
    \right)
    \quad \text{and} \quad
    \tau_\Theta
    \coloneqq
    \left(
        \tau_\Theta^i,
        \tau_\Theta^{-i}
    \right).
\]

\noindent Introducing a multiplier~$\lambda_{v,\theta}^i$ for each residual $\delta_{v,\theta}^i$ yields the augmented Lagrangian
\begin{equation}
\label{eq:cascaded_global_lagrangian}
    \mathcal L_\rho^i \left(
    \tau_\Theta,
    c_{\mathcal V}^i,
    \lambda_\Theta^i
\right)
    \coloneqq
    \sum_{\theta\in\Theta}
    \mathcal L_{\rho,\theta}^i\left(
        \tau_\theta,
        c_{\mathcal V(\theta)}^i,
        \lambda_\theta^i
    \right) ,
\end{equation}
where the scenario-local augmented Lagrangian is
\begin{equation}
\label{eq:cascaded_local_lagrangian}
\begin{aligned}
\mathcal L_{\rho,\theta}^i&
\coloneqq{}
\eqnmarkbox[olive]{cost}{
    b(\theta)J^i(\tau_\theta)
}%
\rlap{%
    \hspace{0.15em}%
    \textcolor{olive}{%
        \raisebox{0.45ex}{\rule{0.8em}{0.4pt}}%
    }%
    \hspace{0.4em}%
    \textcolor{olive!85}{%
        \sffamily\footnotesize
        expected-cost contribution%
    }%
}
\\
&+
\eqnmarkbox[purple]{augmented}{\sum_{v\in\mathcal V(\theta)}
\left[
(\lambda_{v,\theta}^i)^\top\delta_{v,\theta}^i
+
\frac{\rho}{2}\|\delta_{v,\theta}^i\|_2^2
\right]}.
\end{aligned}
\end{equation}

\annotate[yshift=-0.4em]{below,right}
    {augmented}{augmented terms}

\medskip
\noindent Eq.~\eqref{eq:cascaded_global_lagrangian} only weights by belief the scenario's contribution to expected cost. Each local augmented Lagrangian contains one consensus term for every node $v\in\mathcal V(\theta)$ along scenario $\theta$'s tree path.

\subsection{Parallel scenario update}
\label{sec:parallel_scenario_update}

At ADMM iteration~$k$, the consensus prefixes and multipliers are held
fixed. We use the superscript $k$ to denote the ADMM iteration. Eq. (\ref{eq:scenario_update}) gives the scenario update for the ego component. For each $\theta$ in $\Theta$, all agents simultaneously solve their optimization problems. The non-ego agents keep their original problems while the robot's problem considers the augmented Lagrangian as objective \eqref{eq:cascaded_local_lagrangian}.
\begin{subequations}
\label{eq:scenario_update}
\begin{align}
    \tau_\Theta^{i,k+1}
    \in
    \argmin_{\tau_\Theta^i}
    \quad&
    \mathcal L_{\rho}^i
    \left(
        \tau_\Theta,
        c_{\mathcal V}^{i,k},
        \lambda_\Theta^{i,k}
    \right)
    \label{eq:scenario_update_cost}
    \\
    \textup{subject to}
    \quad&
    K^i\!\left(
        \tau_\theta
    \right)\geq 0,
    \qquad
    \forall\theta\in\Theta.
    \label{eq:scenario_update_constraint}
\end{align}
\end{subequations}

Computationally, we impose  the agents' first-order KKT conditions and assemble the system into a scenario-specific Mixed Complementarity Problem (MCP). Since the consensus variables and multipliers are fixed, the MCP for scenario $\theta$ depends only on the trajectory profile
$\tau_\theta$. The MCPs associated with different scenarios therefore share no decision variables and can be solved in parallel.

\subsection{Node-dependent consensus update}
\label{sec:node_consensus_update}

Following the scenario update at iteration $k$, the scenario-dependent trajectories are reconciled independently before each node-wise branching time. After the trajectory update, we formulate the node-wise consensus residual as a function of the common prefix candidate for node $v$ and for $\theta \in \Theta_v$
\begin{equation}
\label{eq:cascaded_disparity}
    \delta_{v,\theta}^{i,k+1}(c_v^{i}, \tau_\theta^{k+1})
    =
    P^u_v\tau_\theta^{i,k+1}
    -
    c_v^{i}.
\end{equation}

\noindent Holding the updated trajectories and multipliers fixed, the consensus prefix associated with node~$v$ is obtained by minimizing the augmented Lagrangian in (\ref{eq:cascaded_local_lagrangian}) with respect to the consensus trajectory $c_v^i$. For every node $v\in\mathcal V$, minimizing with respect to $c_v^i$ gives the closed-form update

\begin{equation}
\label{eq:conditional_consensus_update}
c_v^{i,k+1}
=
\frac{1}{|\Theta_v|}
\sum_{\theta\in\Theta_v}
\left(
    P_v^u\tau_\theta^{i,k+1}
    +
    \frac{\lambda_{v,\theta}^{i,k}}{\rho}
\right).
\end{equation}

\noindent Update~\eqref{eq:conditional_consensus_update} is the main difference from single-trunk consensus splitting. Rather than averaging all scenario-dependent trajectories into a single common prefix, it computes a separate consensus prefix for every branching node in parallel. 

\subsection{Multiplier update}
\label{sec:multiplier_reconstruction}

Following the standard unscaled ADMM dual ascent step
\cite{boyd:2011},  each multiplier is updated by adding $\rho$ times the primal residual associated with its consensus constraint in \eqref{eq:tree_consensus}

\begin{equation}
\label{eq:cascaded_multiplier_update}
    \lambda_{v,\theta}^{i,k+1}
    =
    \lambda_{v,\theta}^{i,k}
    +
    \rho
    \eqnmarkbox[purple]{disagreement}{\left(
        P^u_v\tau_\theta^{i,k+1}
        -
        c_v^{i,k+1}
    \right)}.
\end{equation}
\annotate[yshift=-0.5em]{below, right}{disagreement}{consensus residual}
\smallskip

\noindent Let $r^k$ and $s^k$ respectively denote the primal and dual consensus residuals. The algorithm terminates when
\[
    \lVert r^k\rVert_2\leq\varepsilon_{\mathrm{pri}},
    \qquad
    \lVert s^k\rVert_2\leq\varepsilon_{\mathrm{dual}},
\]
or when the maximum iteration count is reached. Since the trajectory games are nonconvex, standard convex ADMM guarantees do not apply. However, provided that the iterates remain in a regular neighborhood, that the quadratic penalization parameter $\rho$ is sufficiently large, and the scenario subproblems are solved with summable errors, local nonconvex ADMM results \cite{hong:2015} imply vanishing consensus residuals and stationarity of accumulation points. However, the local guarantee neither ensures convergence from arbitrary initializations nor establishes that the resulting stationary point is a GNE. We therefore report the returned solution as a stationary equilibrium candidate whose degree of consensus is assessed through $r^k$ and $s^k$.

\section{Empirical analysis}

In this section, we conduct experiments whose results illustrate that the multi-branching method improves performance when several sources of intent uncertainty are resolved asynchronously. Furthermore, we show that the ADMM-enabled multi-branching solver runs more rapidly in receding-horizon that its single branching time counterpart \cite{peters:2024}. We also present a proof of concept for solving interactions in a different, more complex geometry than the three-agent scenario.

\subsection{Closed-loop simulation framework}
\label{sec:experimental_environment}
At each simulation step, we estimate a sequence of ordered branching times from the current belief distribution. Given this structure, we then solve 
a contingency game over a planning horizon $T_{\mathrm{plan}}$, using the current, fixed belief distribution and the ordered set of branching times. The robot applies the first control of his root control prefix while non-ego agents apply the control obtained from solving their ground-truth intent optimization problem. This game transition yields new Gaussian noise-corrupted observations about the uncertain agents, which are used to update both the belief and the branching times. This procedure is repeated until the episode terminates because the robot reaches its goal, a collision happens, or the simulation horizon $T_{\mathrm{sim}}$ is exceeded.

\smallskip
\noindent
\textsc{belief update} \quad
The joint belief over intent combinations is updated between planner calls using Bayes' rule, by comparing the noisy observations with the states predicted under each intent combination, as in \cite{peters:2024}. The observation likelihood is represented by a Gaussian mixture model, which evaluates how likely the observed non-ego state is under the state prediction associated with each scenario $\theta \in \Theta$. 

\smallskip
\noindent
\textsc{estimation of branching times} \quad
We extend the belief-propagation estimator of Peters et al.~\cite{peters:2024} to each node. Consider node $v\in\mathcal V_q$ at simulation time $t$. Let $\widehat b_{v,t+\ell}^q(\cdot\mid\theta)$ denote the marginal belief over agent $q$'s intents, conditioned on the intent history $h_v$, predicted $\ell$ steps ahead by propagating hypothetical observations under scenario $\theta$. The corresponding branching time is obtained by solving

\begin{equation}
\label{eq:node_branching_time}
\begin{aligned}
t_v
=
t+\;\max_{\theta\in\Theta_v} \;
\min&_{\ell\in\{1,\ldots,T_{\mathrm{plan}}-1\}}\ \quad \ell \\
\; \text{subject to}& \quad
\bar H\!\left(
\widehat b_{v,t+\ell}^{q}(\cdot\mid\theta)
\right)
\leq\varepsilon_H,
\end{aligned}
\end{equation}
\noindent with $\bar H$ the normalized entropy and $\varepsilon_H\in[0,1]$ the prescribed resolution threshold. The normalized entropy measures the remaining uncertainty about agent $q$'s intent.

For each possible scenario $\theta$, compatible with node $v$, we compute the earliest time at which agent $q$'s intent is estimated to be sufficiently resolved, and select the latest of these times. The joint belief is conditioned on the corresponding resolved intent, normalized by the probability mass of the parent branch, and used to estimate the branching time for the next uncertain agent. Repeating this construction recursively produces node-dependent branching times.

\begin{figure}[!t]
    \centering
    \begingroup
        \sffamily
        \def\svgwidth{0.9\columnwidth}
        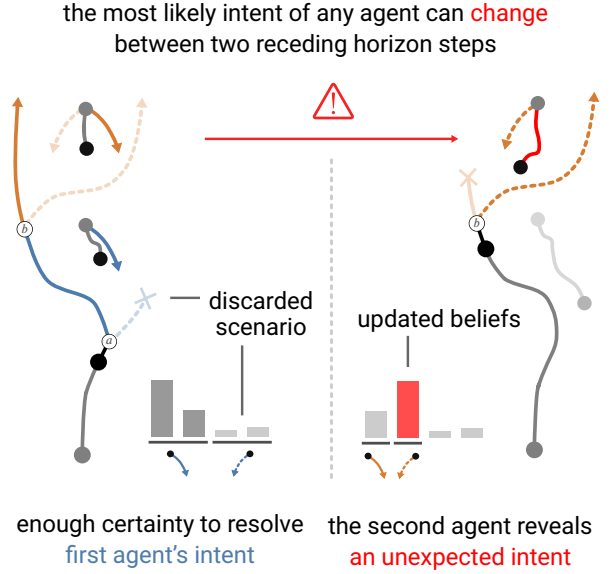
      \endgroup
    \caption{Overview of a receding-horizon simulation. At each time step, the robot executes the first action of the open-loop solution with the current belief and branching times. When the next branching time is imminent, the ego picks one branch and discards the others.}
    \label{fig:receding_fig}
\end{figure}

\subsection{Numerical setup}

We implement the proposed method in Julia within \contingencygames~\cite{contingencygames}. The local MCPs are solved concurrently using \mixedmcp~\cite{fridovichkeil:2026}. In the reported experiments, the consensus residuals are scaled by $\sqrt{b(\theta)}$, replacing the uniform average in \eqref{eq:conditional_consensus_update} with the conditional weights $b(\theta)/\beta_v$ to allow the iterates to reflect the current belief. This implementation-level preconditioning is not used in the theoretical development. The experiments are performed on an Apple silicon processor with eight performance cores, using eight Julia threads. The generated MCP representations and solver structures are reused throughout the study to amortize compilation costs. Within each closed-loop episode, the previous controls and primal variables are shifted forward by one step. Terminal controls and ADMM multipliers are initialized to zero, and the states are regenerated through the dynamics. Symbolic construction, compilation, and the initial cold-started solve are excluded from the reported online computation time. Therefore, the reported solve time corresponds to the empirical mean numerical solve time per simulation step.

\subsection{Experimental protocol}

We extend the collision-avoidance scenario of Peters et al.~\cite{peters:2024}, involving a robot and a human agent crossing a road, by adding a second human agent behind the first one with the same two possible intents (cf. Fig.~\ref{fig:receding_fig}). The motion of the second human is delayed, making both the interactions and the informational gains with respect to both human agents asynchronous.

We conduct a Monte Carlo study over $100$ paired receding-horizon simulations drawn from different variations of the three-agent scenario. The four intent combinations are represented in equal proportion across the 100 simulations. To vary the temporal separation between the two uncertain agents, we uniformly sample the initial lateral positions $(x_0^1,x_0^2)$, the first agent's initial distance to the robot \(\Delta y_1\), the inter-agent distance \(d_{12}\) and the delay in the second agent's motion \(\tau_2\) following
\begin{equation*}
\begin{aligned}
x_0^1,& x_0^2 \sim \mathcal U[-0.10,0.10],
&\qquad
\Delta& y_1 \sim \mathcal U[-0.10,0.10], \\
d_{12} &\sim \mathcal U[1.35,1.85],
&\qquad
\tau_2 &\sim \mathcal U\{6,\ldots,10\}\ \text{steps},
\end{aligned}
\end{equation*}
where $\mathcal U(\cdot)$ denotes the uniform distribution over the associated set. All distances are given in arbitrary units. The remaining simulation parameters are fixed across all trials.\footnote{Monte Carlo settings: $T_{\mathrm{plan}}=36$, $T_{\mathrm{sim}}=30$, $\Delta t=0.2\,\mathrm{s}$, $\sigma_{\mathrm{obs}}=0.04$, $d_{\mathrm{safe}}=0.85$, $\rho=50$, and eight ADMM iterations per planning step for both formulations.} Each sampled instance is solved once under each branching structure receiving identical initial conditions, ground-truth intents, observation noise, and random seed.

Interaction success requires the ego to pass both agents within the simulation horizon without safety violations. We assess progress efficiency via the final cost per executed step over the receding-horizon simulation. To avoid conflating interaction success and trajectory quality, efficiency and comfort are compared only over runs successfully completed by both methods. Comfort is evaluated by the root mean square (RMS) variation of consecutive normalized ego controls actually applied to the ego over $N_{exec}$ steps,

\begin{equation}
S =
\sqrt{
\frac{1}{N_{\mathrm{exec}}-1}
\sum_{k=2}^{N_{\mathrm{exec}}}
\left\|
\frac{{u^i}_{k}
-
{u^i}_{k-1}}{u_{\mathrm{max}}}
\right\|_{2}^{2}
},
\end{equation}
\noindent where $u_{\max}$ is the maximum admissible control magnitude.

Lower values indicate less sudden variations and therefore smoother trajectories. Computational performance is measured using the mean online solve time per simulation step, excluding symbolic problem building and warm-start. For every metric, we report empirical performance with $95\%$ confidence intervals.

\section{Results}

\subsection{Monte Carlo branching structure comparison}

\begin{figure}[!t]
    \centering
    \vspace*{10pt}
    \begingroup
        \sffamily
        \def\svgwidth{0.94\columnwidth}
        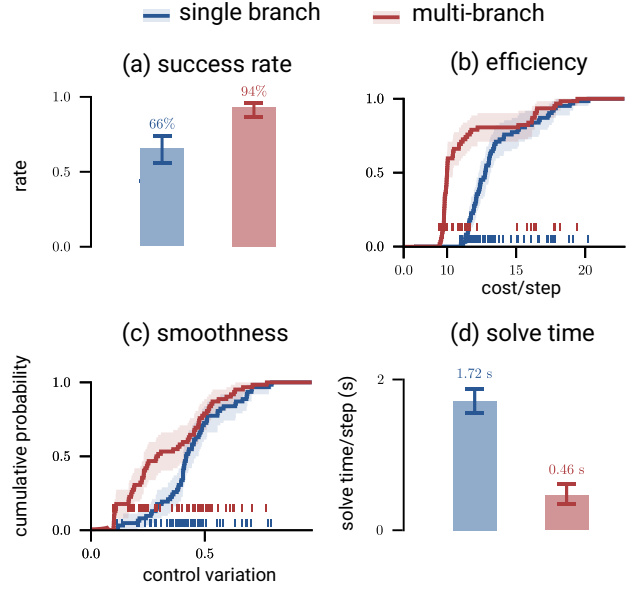
    \endgroup
    \caption{Monte Carlo study results for the three-agent contingency game.}
    \label{figure:montecarlo_results}
\end{figure}

The ability to progressively adapt the shared controls as uncertainty is resolved improves progress efficiency. The multi-branch formulation increases  the success rate from $66\%$ to $94\%$ compared to the single-branch structure~(Fig.~\ref{figure:montecarlo_results}a). Among the $62$ runs successfully completed by both methods, the empirical distribution of the average cost per step is shifted towards lower values with the multi-branch formulation (Fig.~\ref{figure:montecarlo_results}b). 

A similar shift is observed for the RMS control variation~(Fig.~\ref{figure:montecarlo_results}c). The multi-branch formulation produces trajectories displaying lower control variations across successful cases. These results suggest that multi-branch structure better anticipates the asynchronous revelation of the agents' intent.

These improvements in performance and control smoothness do not come at the expense of computational efficiency in the considered implementation (Fig.~\ref{figure:montecarlo_results}). The mean solve time per planning step decreases from \(1.72\,\mathrm{s}\) with the single-branch formulation to \(0.46\,\mathrm{s}\) with the multi-branch formulation, representing a reduction of approximately \(73\%\). Moreover, the multi-branch method is faster in 97$\%$ of the cases, including the $62$ shared successful cases. Despite its more complex structure, the proposed method achieved a mean sub-second receding-horizon solve time in the considered experiments.

\subsection{Qualitative failure analysis}

\begin{figure}[!b]
    \centering
    \begingroup
        \sffamily
        \def\svgwidth{\columnwidth}
        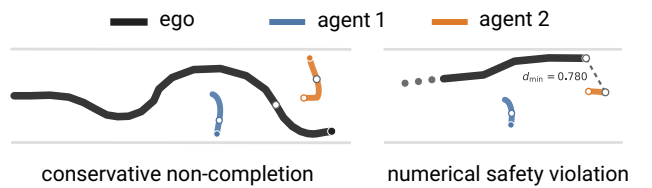
    \endgroup
    \caption{Failure patterns illustration from the Monte Carlo study.}
    \label{figure:failure_patterns}
\end{figure}

The observed failures arise when the two interactions are temporally and spatially coupled (Fig.~\ref{figure:failure_patterns}). Close spacing between uncertain agents causes the second interaction to happen shortly after the first. Delay in the second agent's relevant motion leaves less time for the robot to react. This difficulty is exacerbated when both uncertain agents have opposing intents requiring better anticipation. By contrast, the lateral initial positions and the distance to the first agent exhibit reduced correlation to failure. These results show that the adaptation window between successive interactions must be sufficiently large to allow the robot to effectively react.

Two distinct failure modes emerge in this study. The first is due to overconservative ego behavior, marked by a lack of progress due to excessive slowdowns near the collision boundary. The robot stops because its previous controls did not sufficiently anticipate the second interaction, leaving it with no feasible trajectories to pass the interaction. The second mode is associated with a solver failure caused by insufficient ADMM convergence near the feasibility boundaries. In those cases, the convergence issue leads to safety violations. Note that, however, the existing single branching time formulation of \cite{peters:2024} also fails in most of these difficult cases: single branching either makes the robot commit while not having sufficient certainty about the second interaction, or produces an overly conservative trajectory. Nevertheless, multi-branching does not strictly dominate in every simulated interaction.

\subsection{Overtaking scenario analysis}

\begin{figure}[!t]
    \centering
    \vspace*{10pt}
    \begingroup
        \sffamily
        \def\svgwidth{0.90\columnwidth}
        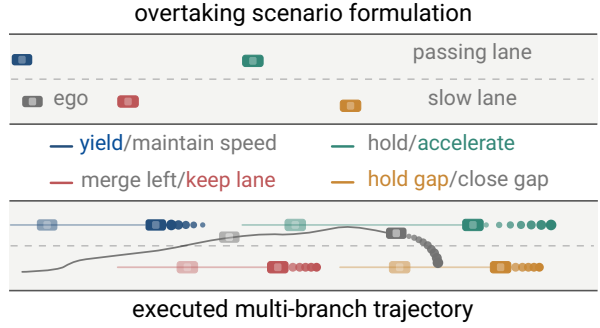
    \endgroup
    \caption{Collision-avoidance problem for an overtaking scenario. For each vehicle, we show the two intents and the selected one in color. The multi-branch method safely overtakes the slow agents while avoiding others.}
    \label{figure:paper_overtaking}
\end{figure}

We also demonstrate the method in a larger-scale highway-overtaking interaction with four uncertain traffic agents, each having two possible intents, yielding $16$ joint intent scenarios (Fig.~\ref{figure:paper_overtaking}). We use branching times $(5,7,9,11)$ and a budget of $25$ ADMM iterations per replanning step to limit online computation. With this budget, the single-branch formulation fails to complete the maneuver within the simulation horizon. By contrast, The multi-branch ego overtakes the slower vehicle while avoiding other agents. The maneuver terminates safely after $58$ simulation steps with a median online solve time of $2.14\,\mathrm{s}$.

\section{Conclusion}

Multi-branch contingency games model interactions between a robot and agents whose intents are resolved asynchronously. We develop an ADMM scenario decomposition that follows a logical information tree, enabling partial consensus that reflects the branching structure that describes when intent hypotheses are resolved. Via decomposition and parallel computation, we can solve games with more branches and reach an average online solve time of \(0.46\,\mathrm{s}\) per receding-horizon step across $100$ Monte Carlo cases. The multi-branch method increases the overall success rate by $28$ points compared to the existing single branching time formulation in \cite{peters:2024}, while reducing the ego's cost by $15\%$ and control variation by $26\%$ among simulations both methods complete successfully. We also demonstrate that the proposed approach successfully scales to a five-agent overtaking scenario with 16 considered intent scenarios distributed across four uncertain agents. These results show that hierarchical contingency structures can help robots plan more efficiently in uncertain multi-agents interactions. Ultimately, future work could relax the assumption of finite intent spaces through continuous intent representations.

\section{Acknowlegements}
B. Lechardoy, P. de las Heras Molins, L. Pautet, and G. Bakirtzis are partially supported by the academic and research chair
\emph{Architecture des Systèmes Complexes} through the following partners: Dassault Aviation, Naval
Group, Dassault Systèmes, KNDS France, Agence de l'Innovation de Défense, and
Institut Polytechnique de Paris.

D. Fridovich-Keil is partially supported by the Hi! PARIS international visiting chair program.
\printbibliography
\end{document}